\PassOptionsToPackage{dvipsnames}{xcolor}
\documentclass{article}

\usepackage{microtype}
\usepackage{graphicx}
\usepackage{subfigure}
\usepackage{booktabs} %

\usepackage{hyperref}

\usepackage[accepted]{icml2025}

\usepackage{amsmath}
\usepackage{amssymb}
\usepackage{mathtools}
\usepackage{amsthm}

\usepackage[capitalize,noabbrev]{cleveref}

\theoremstyle{plain}

\theoremstyle{definition}

\theoremstyle{remark}

\usepackage[textsize=tiny]{todonotes}

\usepackage{amsmath}
\usepackage{amssymb}
\usepackage{bm}
\usepackage{mathtools}

\def\vtheta{{\bm{\theta}}}

\def\vdelta{{\bm{\delta}}}
\def\vlambda{{\bm{\lambda}}}

\def\va{{\bm{a}}}
\def\vb{{\bm{b}}}

\def\ve{{\bm{e}}}

\def\vf{{\bm{f}}}
\def\vg{{\bm{g}}}

\def\vv{{\bm{v}}}

\def\vx{{\bm{x}}}
\def\vy{{\bm{y}}}

\def\vz{{\bm{z}}}

\def\mA{{\bm{A}}}
\def\mB{{\bm{B}}}
\def\mC{{\bm{C}}}

\def\mE{{\bm{E}}}
\def\mF{{\bm{F}}}
\def\mG{{\bm{G}}}

\def\mH{{\bm{H}}}
\def\mI{{\bm{I}}}

\def\mQ{{\bm{Q}}}

\def\mS{{\bm{S}}}

\def\mU{{\bm{U}}}

\def\mW{{\bm{W}}}
\def\mX{{\bm{X}}}

\def\mZ{{\bm{Z}}}

\DeclareMathAlphabet{\mathsfit}{\encodingdefault}{\sfdefault}{m}{sl}
\SetMathAlphabet{\mathsfit}{bold}{\encodingdefault}{\sfdefault}{bx}{n}

\def\gA{{\mathcal{A}}}

\def\gI{{\mathcal{I}}}

\def\gL{{\mathcal{L}}}

\def\gN{{\mathcal{N}}}

\def\sB{{\mathbb{B}}}

\def\sD{{\mathbb{D}}}

\def\sR{{\mathbb{R}}}

\DeclareSymbolFont{bbold}{U}{bbold}{m}{n}
\DeclareSymbolFontAlphabet{\mathbbold}{bbold}

\newcommand{\E}{\mathbb{E}}

\DeclareMathOperator*{\argmin}{arg\,min}

\DeclareMathOperator{\Tr}{Tr}
\DeclareMathOperator{\diag}{diag}

\DeclareMathOperator{\Cat}{Cat}
\DeclareMathOperator{\flatten}{vec}

\DeclareMathOperator{\bigO}{\mathcal{O}}
\usepackage{comment} %

\ifdefined\isaccepted
  \newcommand{\linkToDocs}{https://curvlinops.readthedocs.io}
  \newcommand{\linkToGithub}{https://github.com/f-dangel/curvlinops}
\else
  \newcommand{\linkToDocs}{anonymized}
  \newcommand{\linkToGithub}{anonymized}
\fi

\usepackage{todonotes}
\usepackage{xspace}
\newcommand*{\ie}{i.e.\@\xspace}
\newcommand*{\iid}{i.i.d.\@\xspace}
\newcommand*{\wrt}{w.r.t.\@\xspace}
\newcommand*{\eg}{e.g.\@\xspace}

\newcommand*{\Eg}{E.g.\@\xspace}

\crefname{section}{\S\!\!}{\S\!\!}
\crefname{appendix}{\S\!\!}{\S\!\!}

\Crefname{equation}{Eq.}{Eqs.}
\Crefname{figure}{Fig.}{Figs.}
\Crefname{tabular}{Tab.}{Tabs.}

\usepackage{nicefrac} %

\usepackage{multirow}

\usepackage{enumitem} %

\definecolor{VectorBlack}{RGB}{34, 34, 34}
\definecolor{VectorGray}{RGB}{239, 238, 237}

\definecolor{VectorBlue}{RGB}{59, 69, 227}
\definecolor{VectorPink}{RGB}{253, 8, 238}
\definecolor{VectorOrange}{RGB}{250, 173, 26}
\definecolor{VectorTeal}{RGB}{82, 199, 222}

\usepackage{listings}
\lstdefinestyle{vector_institute}{
  backgroundcolor=\color{VectorGray!50},
  commentstyle=\bfseries\rmfamily\color{VectorBlue},
  columns=fullflexible,
  keywordstyle=\bfseries\color{VectorBlack},
  numberstyle=\tiny\color{VectorBlack!50},
  stringstyle=\bfseries\color{VectorPink},
  basicstyle=\ttfamily\small,
  xleftmargin=3.2ex,
  breakatwhitespace=false,
  breaklines=true,
  captionpos=t,
  keepspaces=true,
  numbers=left,
  numbersep=7pt,
  showspaces=false,
  showstringspaces=false,
  showtabs=false,
  tabsize=2,
  escapebegin={\color{blue}},
  linewidth=\linewidth,
  mathescape=true,
}
\usepackage{tikz}
\usetikzlibrary{arrows.meta}

\usepackage{soul}

\usepackage[leftcaption]{sidecap}
\sidecaptionvpos{figure}{t}
\usepackage{graphicx}
\usepackage[export]{adjustbox}

\begin{document}

\newcommand{\papertitle}{%
  Position: Curvature Matrices Should Be Democratized via Linear Operators
}

\twocolumn[
  \icmltitle{\papertitle}

  \icmlsetsymbol{equal}{*}

  \begin{icmlauthorlist}
    \icmlauthor{Felix Dangel}{equal,vector}
    \icmlauthor{Runa Eschenhagen}{equal,cambridge}
    \icmlauthor{Weronika Ormaniec}{eth}
    \icmlauthor{Andres Fernandez}{tuebingen}
    \icmlauthor{Lukas Tatzel}{tuebingen}
    \icmlauthor{Agustinus Kristiadi}{vector}
  \end{icmlauthorlist}

  \icmlaffiliation{vector}{Vector Institute, Toronto, Canada}
  \icmlaffiliation{eth}{ETH, Z\"urich, Switzerland}
  \icmlaffiliation{tuebingen}{T\"ubingen AI Center, University of T\"ubingen, T\"ubingen, Germany}
  \icmlaffiliation{cambridge}{Cambridge University, United Kingdom}

  \icmlcorrespondingauthor{Felix Dangel}{fdangel@vectorinstitute.ai}

  \icmlkeywords{PyTorch, Curvature matrices, Hessian, Fisher, KFAC, Generalized Gauss-Newton, Linear operators, Large-scale linear algebra}

  \vskip 0.3in
]

\printAffiliationsAndNotice{\icmlEqualContribution} %

\begin{abstract}
    Structured large matrices are prevalent in machine learning. 
    A particularly important class is curvature matrices like the Hessian, which are central to understanding the loss landscape of neural nets (NNs), and enable second-order optimization, uncertainty quantification, model pruning, data attribution, and more. 
    However, curvature computations can be challenging due to the complexity of automatic differentiation, and the variety and structural assumptions of curvature proxies, like sparsity and Kronecker factorization.
    In this \textbf{position paper}, we argue that \textbf{linear operators---an interface for performing matrix-vector products---provide a general, scalable, and user-friendly abstraction to handle curvature matrices}. 
    To support this position, we developed \texttt{curvlinops}, a library that provides curvature matrices through a unified linear operator interface. 
    We demonstrate with \texttt{curvlinops} how this interface can hide complexity, simplify applications, be extensible and interoperable with other libraries, and scale to large NNs.
\end{abstract}

\section{Introduction}
\paragraph{Structured matrices in ML.}
Large matrices that exhibit structure play an important role in various machine learning (ML) applications:
\emph{Sparse} matrices contain only a small fraction of non-zero entries, and are, \eg, common in learning problems on graphs \cite{kipf2017semisupervised} whose adjacency matrices are sparse.
\emph{Symbolic} matrices \cite{charlier2021kernel} are matrices whose coefficients are specified through a mathematical expression (say $\lVert \vx_i - \vx_j \rVert_2$ or $k(\vx_i, \vx_j)$) and include kernel matrices used in Gaussian Processes~\cite{williams2006gaussian}, distance matrices used for k-means clustering~\cite{steinhaus1956division} or k-NN classification/regression~\cite{fix1951discriminatory}, and attention \cite{vaswani2017attention} matrices.
\emph{Factorized} matrices consist of one or multiple low-dimensional, and therefore parameter-efficient, tensor components that are combined via tensor multiplications. Examples include
diagonal, block-diagonal, Kronecker \cite{loan2000ubiquitous}, outer product \citep[or low rank,][]{jacot2020neural,ren2019efficient,dangel2022vivit}, (block) tensor-train, and Monarch \cite{dao2022monarch} matrices, but can also be generalized to unnamed factorizations using tensor networks~\cite{potapczynski2024searching,penrose1971applications,bridgeman2017hand,biamonte2017tensor}.
Other `low-complexity' structures are butterfly \cite{dao2019learning}, Toeplitz (or circulant), low-rank plus diagonal \cite{lin2024structured}, or hierarchical matrices~\cite{chen2022efficient}.
These structures allow executing operations like matrix-vector products \emph{matrix-free}, \ie without materializing the dense matrix.

\definecolor{color1}{HTML}{B8C4BB}
\definecolor{color2}{HTML}{663F46}
\definecolor{color3}{HTML}{3C362A}
\definecolor{color4}{HTML}{C9D6EA}

\begin{SCfigure*}[0.41] %
\begin{wide} 
\centering
\begin{adjustbox}{width=1.1\linewidth}
\begin{tikzpicture}
  \tikzstyle{every node}=[semithick,font=\footnotesize]
  
    \tikzstyle{mystyle}=[draw, thick, rounded corners, align=center]

    \node[mystyle, name=input, fill=color3!20] {
      \textbf{Empirical risk}
      \\[1ex]
      $\sD, f_\vtheta, \ell$
      \\[1ex]
      (data, net, loss)
    };
    \node[mystyle, name=linops, right=1.25cm of input, fill=color1] {
      \textbf{Linear operator}
      \\[1ex]
      $\vv \mapsto \gA(\vv)$
      \\[1ex]
      (options: \Cref{fig:visual-tour})
    };
    \node[mystyle, name=highops, below=of linops, fill=color1] {
      \textbf{Transformation}
      \\[1ex]
      $\gA \mapsto f(\gA)$
      \\[1ex]
      (e.g.\,inverse)
    };
    \node[mystyle, name=scipy, below=of highops, fill=color2!50] {
      \textbf{SciPy export}
      \\[1ex]
      \texttt{.to\_scipy()}
      \\[1ex]
      (compatibility)
    };
    \node[mystyle, name=properties, right=1.25cm of linops, yshift=-8.5ex, fill=color1] {
      \textbf{Estimation}
      \\[1ex]
      $\Tr(\cdot), \diag(\cdot), \lVert\cdot\rVert_{\text{F}}^2, \rho(\cdot)$
      \\[1ex]
      (e.g.\,Hutchinson)
    };
    \node[mystyle, name=scipy_prop, below=5.25ex of properties, fill=color2!50] {
      \textbf{Use SciPy}
      \\[1ex]
      \texttt{estimate\_rank},\\
      \texttt{svds}, \texttt{eigsh}
      \\[1ex]
      (well-established)
    };
    \node[mystyle, name=applications, right=1.25cm of properties, fill=color4, align=left, yshift=-3.5em] {
      \qquad\,\,\,\textbf{Applications}
      \\[1ex]
      • 2\textsuperscript{nd}-order optimization,
      \\
      \phantom{•} natural gradient descent
      \\[0.75ex]
      • Model merging (Fisher)
      \\[0.75ex]
      • Training data attribution
      \\
      \phantom{•} (influence functions)
      \\[0.75ex]
      • Bi-level optimization
      \\
      \phantom{•} (e.g.\,adversarial attacks)
      \\[0.75ex]
      • Model sparsification
      \\[0.75ex]
      • Loss landscape analysis
      \\
      \phantom{•} (e.g.\,Hessian spectrum)
      \\[0.75ex]
      • Uncertainty quantification
      \\
      \phantom{•} (Laplace approximations)
    };

    \tikzset{>={Stealth}}
    \draw[->, ultra thick] (input) to (linops);
    \draw[->, ultra thick] (linops) -- (highops);
    \draw[->, ultra thick] (linops.south west) to[in=135, out=225] (scipy.north west);
    \draw[->, ultra thick] (linops.south east) to[out=315, in=180] (properties.west);
    \draw[->, ultra thick] (highops.north east) to[out=45, in=180] (properties.west);
    \draw[->, ultra thick] (highops) -- (scipy);
    \draw[->, ultra thick] (properties) to (applications.west |- properties);
    \draw[->, ultra thick] (scipy.north east) to[out=45, in=180] (scipy_prop);
    \draw[->, ultra thick] (scipy_prop) to (applications.west |- scipy_prop);
    \draw[->, ultra thick] (scipy) to (applications.west |- scipy);
    \draw[->, ultra thick] (highops) to (applications.west |- highops);
    \draw[->, ultra thick] (linops) to (applications.west |- linops);

    \node [anchor = south west, fill=black!5, rounded corners] at (current bounding box.south west) {\includegraphics[scale=1.1]{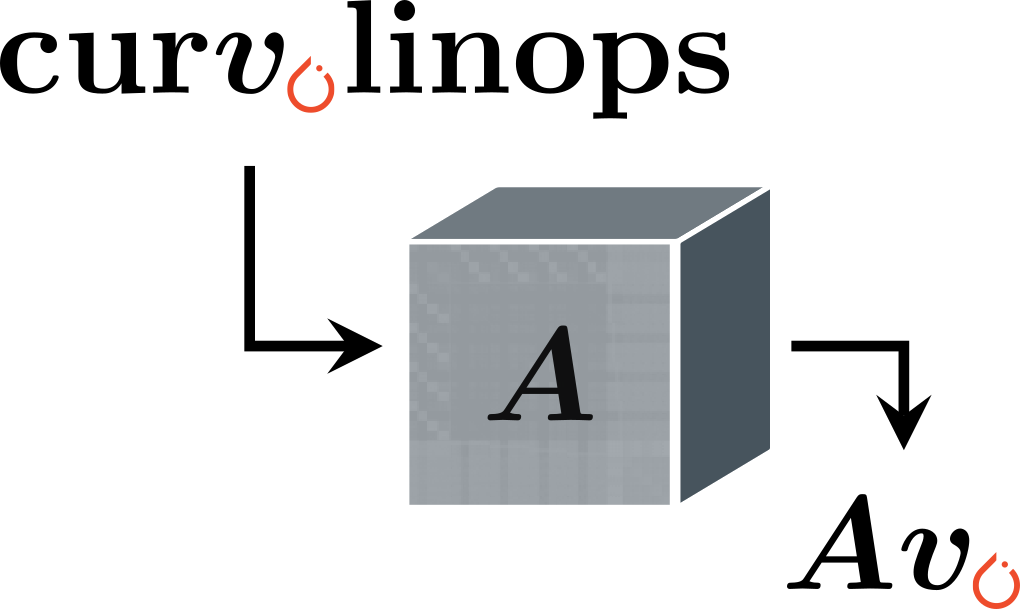}};
  \end{tikzpicture}
\end{adjustbox}
  \caption{\textbf{Overview of \texttt{curvlinops}'s features.}
    At its core, the library provides linear operators for curvature matrices of an empirical risk in PyTorch.
    Operator transformations allow representing matrix functions like inversion.
    To analyze the underlying matrices, we implement known estimation techniques based on randomized linear algebra.
    Other routines are already provided in SciPy, and they become accessible through \texttt{curvlinops}'s export feature.
  }
  \label{fig:overview}
\end{wide}
\end{SCfigure*}
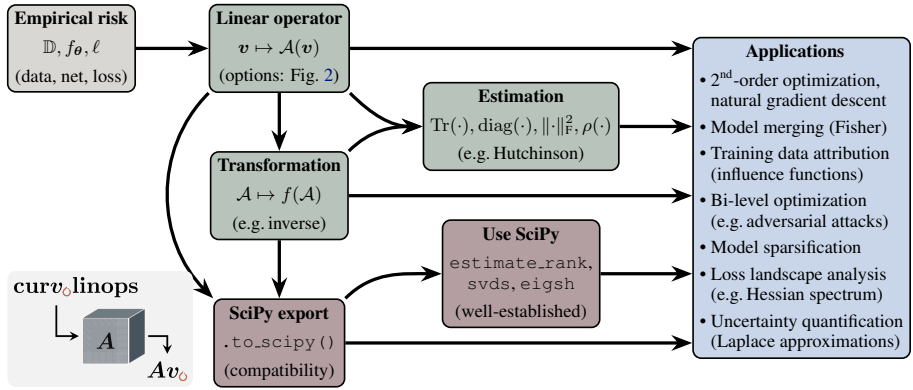

\paragraph{Linear operators.}
In many scenarios, it suffices to query matrix-vector products, rather than access a dense matrix representation.
Matrix-free structures are useful to accelerate such matrix-vector products for solving, or approximately solving, numerical tasks and to reduce memory demands.
This motivates \emph{linear operators}, an abstraction for carrying out black-box matrix-vector multiplications. %
More formally, a linear operator $\gA$ is a linear map between two vector spaces, $\gA: \sR^{D_{\text{in}}} \to \sR^{D_{\text{out}}}, \,\vv \mapsto \gA(\vv)$, such that for $\vv_1, \vv_2 \in \sR^{D_{\text{in}}}$ and $\alpha \in \sR$,
\begin{align*}
  \gA(\vv_1 + \vv_2) & = \gA(\vv_1) + \gA(\vv_2)\,,
  \\
  \gA(\alpha \vv)    & = \alpha \gA(\vv)\,.
\end{align*}
We can think of $\gA$ as \emph{action} of a matrix $\mA \in \sR^{D_{\text{out}} \times D_{\text{in}}}$ with entries $[\mA]_{i,j} = \ve_i^{\top} \gA(\ve_j)$ with $\ve_i$ the $i$-th standard basis vector.
While $\gA(\vv) = \mA \vv$, the linear operator $\gA$ may not construct the full matrix representation $\mA$, which could be costly.
For example, if $\mA = \diag(\va)$ is a diagonal matrix, the matrix-vector product can be performed via element-wise multiplication, $\gA: \vv \mapsto \mA \vv = \va \odot \vv$, \ie \emph{without} forming $\mA$ in memory.
Since matrix-vector multiplications with a structured matrix can be done more efficiently than explicit multiplication with the dense matrix representation,
linear operators facilitate working with large matrices without the need for explicit storage.

\vspace{20pt}
The fact that structured linear maps can often be implemented more efficiently in both time and memory has spurred the development of iterative algorithms
for estimating the properties of said map, solving linear systems, or computing truncated matrix decompositions from matrix-vector products.
The relevance of these topics for the ML community has been underlined by tutorials at conferences like NeurIPS 2023~\cite{derezinski2024recent}.

\paragraph{Curvature matrices in deep learning.}
Curvature matrices like the Hessian of the empirical risk %
or approximations thereof play a crucial role in describing the loss landscape of neural nets.
They are a necessary ingredient for many applications, e.g.\;second-order optimization \cite{martens2012training,amari2000natural}, uncertainty quantification \cite{mackay1992practical,daxberger2021laplace}, model merging \cite{matena2022merging}, model pruning \cite{lecun1889optimal,singh2020woodfisher}, empirical investigations into the loss landscape \cite{schneider2021cockpit,tatzel2024debiasing,papyan2019spectrum,gurari2018gradient,yao2020pyhessian}, and computing hyper-gradients for training data attribution \cite{koh2017understanding}, adversarial attacks \cite{lu2022indiscriminate}, or hyper-parameter optimization~\cite{lorraine2020optimizing}.

The parameter space of NNs is often so vast that storing even a few columns of these matrices becomes prohibitively expensive.
Additionally, these matrices are often defined on large datasets, necessitating their computation in manageable batches.
Thanks to recent developments of ML libraries~\cite{bradbury2018jax,paszke2019pytorch}, many curvature matrices (e.g.\,the exact Hessian) have become more accessible.
However, matrix-vector products can still be prohibitively costly (equivalents of multiple gradient evaluations~\cite{dagreou2024how}).
As a result, practical curvature matrices often depend on structural approximations to be feasible, e.g.\;Kronecker factorizations~\cite{heskes2000natural,martens2015optimizing}.
These approximations can greatly reduce computational costs while still being useful for applications.
However, implementing and testing these approximations can be extremely challenging, as they come in various forms and use many heuristics.

\begin{figure}[!t]
    \centering
\includegraphics{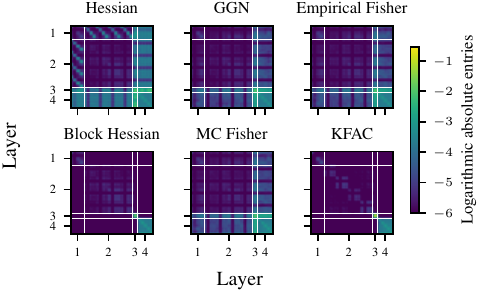}
\vspace{-1ex}
\caption{\textbf{Visual tour of curvature matrices.}
    White lines separate parameters into layers.
    We consider a synthetic classification task with a small convolutional neural net (three convolutional and one dense layer with ReLU and sigmoid activations, $D=683$).
    }
\label{fig:visual-tour}
\vspace{-2.5ex}
\end{figure}

\vspace{20pt}
In this paper, we argue \textbf{our position} that;
\begin{quote}
\textit{Presenting curvature matrices as linear operators empowers users to apply them to various applications without worrying about implementation complexity, and to benefit from their extensibility and interoperability.}
\end{quote}
To substantiate our position, we developed \texttt{curvlinops}%
\ifdefined\isaccepted\footnote{\texttt{pip install curvlinops-for-pytorch}}\fi,
a PyTorch library that offers easy access to various curvature matrices as linear operators. %
We divide our argument in favor of linear operators into three statements and illustrate them through \texttt{curvlinops} (overview in \Cref{fig:overview}):
\begin{enumerate}
    \item \textbf{They encapsulate complexity (\Cref{sec:encapsulation}).}
    We can cover a wide range of curvature matrices---from exact but slow to approximate but fast---in one interface (\Cref{fig:visual-tour}).
    This allows users to focus on their application, liberating them from having to comprehend the numerical details and error-prone implementation intricacies such as correctly scaled accumulation over batches and potential non-deterministic outputs.
    \item \textbf{They simplify applications (\Cref{sec:transferability}).} Because linear operators can be multiplied with vectors like dense matrices, they are easy to use.
    Using code snippets, we show how to implement a diverse set of applications that demand curvature approximations (second-order optimization, model merging, spectral analysis, \dots), demonstrating that the code aligns with the underlying mathematics.
    \item \textbf{They enable extensibility and interoperability (\Cref{sec:interoperability}).}
    We discuss how new curvature approximations can be conveniently implemented as linear operators.
    Moreover, exporting linear operators to SciPy allows the use of its sophisticated algorithmic toolbox for truncated eigendecomposition, SVD, or rank estimation.
    To complement these tools, we implement various (randomized) linear algebra algorithms to estimate matrix properties, such as trace, diagonal, and spectral density.
\end{enumerate}
We also demonstrate scalability to larger architectures like nanoGPT on Shakespeare and ResNet50 on ImageNet (\Cref{sec:performance}).

\paragraph{Related software.}
Various Python packages provide a linear operator interface: PyLops \cite{ravasi2019pylops} focuses on creating and solving inverse problems; the linear operator class in SciPy \cite{virtanen2020scipy} lives in the sparse sub-module.
ML frameworks like PyTorch \cite{paszke2019pytorch} and JAX \cite{bradbury2018jax} do not provide linear operator interfaces, but there are external packages that do so \cite{rader2023lineax,gardner2018gpytorch}.
\citet{potapczynski2023cola,gardner2018gpytorch} additionally incorporate many linear algebra tricks for efficiently solving linear systems, inverting or composing matrices, and computing properties like eigenvalues/spectra.

While there have been efforts to simplify curvature computations~\cite{golmant2018hessian,granziol2019deep,yao2020pyhessian,dangel2020backpack,osawa2023asdl}, achieving support for diverse architectures and extensibility remains challenging.
Our incentive is to unify a wider variety of curvature matrices than previous approaches, and to further disentangle matrix property estimation techniques from matrix-vector products through the linear operator interface.

\section{Linear Operators Encapsulate the Complexity of Curvature Matrices}\label{sec:encapsulation}
Curvature matrices come in various flavors, some of which require knowledge of automatic differentiation (AD), the implementation of loss-specific sampling procedures, or the computation of architecture-specific components such as Kronecker factors.
In addition, computations are usually done in batches and accumulated, which requires adapting the scaling of each result.
Also, the vector in a matrix-vector product can be interpreted as either a single vector or a tensor product.
Lastly, common deep learning elements like data augmentation, dropout~\cite{srivastava2014dropout}, and data-order dependent concepts like batch normalization~\cite{ioffe2015batch} render the represented matrix non-deterministic, which is undesirable.
All these caveats and inconveniences can be encapsulated in a linear operator.

We focus on curvature information of an empirical risk $\gL_{\sD}(\vtheta)$ from a neural net $f_{\vtheta}$ with parameters $\vtheta \in \sR^D$, whose predictions $f_{\vtheta}(\vx)$ on datum $\vx$ are scored on some data set $\sD$ using a convex criterion function $\ell$ and the true label $\vy$,
\begin{align}\label{eq:empirical-risk}
 \textstyle\gL_{\sD}(\vtheta)
  \coloneq
  R_{\sD} \;
  \sum_{\scriptstyle (\vx_n, \vy_n) \in \sD}
  \ell(f_{\vtheta}(\vx_n), \vy_n)\,,
\end{align}
where $R_{\sD}$ is a reduction factor, usually a value from $\{ 1, \nicefrac{1}{|\sD|}, \nicefrac{1}{|\sD| \dim(\vf_n)} \}$, where $\vf_n \coloneq f_{\vtheta}(\vx_n)$.

\subsection{Curvature Matrices}

We provide a quick overview of fundamental curvature matrices like the Hessian, generalized Gauss-Newton (GGN), and different flavors of the Fisher information matrix (Fisher), all of which can be represented using the linear operator interface; see \Cref{fig:visual-tour} for a visual tour.

\paragraph{Hessian.} The empirical risk's Hessian
\begin{align}
  \label{eq:hessian}
  \hspace{-10pt}
  \textstyle\mH(\vtheta)
  \coloneq
  \nabla_{\vtheta}^2 \gL_{\sD}(\vtheta)
  =
  R_{\sD} \;
  \sum_{(\vx_n, \vy_n) \in \sD}
  \nabla_{\vtheta}^2\ell(\vf_n, \vy_n)
\end{align}
collects the second-order derivatives and is thus the most natural choice for the curvature matrix.
While the Hessian is prohibitively large to store, Hessian-vector products (HVPs) with a vector $\vv$ can be executed via nested first-order AD \cite{pearlmutter1994fast}, as $\nabla_\vtheta^2 \gL_\sD(\vtheta) \vv = \nabla_\vtheta ( \nabla_\vtheta \gL_\sD(\vtheta)^\top \vv)$.
\citet{dagreou2024how} empirically probed the efficiency of HVPs and found that they require 2-4x as long as a gradient and consume 2-3x the memory, depending on the usage of just-in-time compilation, and the combination of reverse and forward mode AD. Thanks to the linear operator interface, users are not concerned with the exact HVP implementation.

\paragraph{GGN.}
One issue with the Hessian is that it is usually indefinite.
The GGN $\mG(\vtheta)$ is a positive semi-definite approximation of the Hessian that follows from \emph{partial linearization} of the composition $\ell \circ f_{\vtheta}$ by replacing $f_{\vtheta}$ with a linearization before differentiation,
\begin{align}
  \label{eq:ggn}
  \hspace{-8pt}\mG(\vtheta)
   &\textstyle\coloneqq
  R_{\sD} \;
  \sum_{(\vx_n, \vy_n) \in \sD}
  {\mathrm{J}_{\vtheta}\vf_n}^\top\,
  \nabla_{\vf_n}^2
  \ell(\vf_n, \vy_n)
  \,\mathrm{J}_{\vtheta}\vf_n\,,
\end{align}
with the Jacobian $[\mathrm{J}_{\vtheta}\vf_n]_{i,j} = \nicefrac{\partial [\vf_n]_i}{\partial [\vtheta]_j}$. 
The GGN's positive semi-definite nature is desirable for many applications that demand a convex local quadratic, and GGNVPs can be done with HVPs, VJPs, and JVPs \cite{schraudolph2002fast}.
\citet{martens2010deep} finds that GGNVPs only take about half the memory and run nearly twice as fast as HVPs.
In many tasks, the GGN corresponds to the Fisher~\cite{martens2020new}.

\paragraph{Probabilistic interpretation.}
Empirical risk minimization often amounts to maximizing the likelihood $\hat{p}(\{\vy_n\}_{\sD} \mid \{\vx_n \}_{\sD}, \vtheta) = \prod_{n} p(\vy_n \mid \vx_n, \vtheta)$---or equivalently minimizing the negative log likelihood---where the neural net parameterizes a likelihood $p(\vy \mid \vx, \vtheta) = q(\vy \mid \vf)$ such that $\ell(\vf, \vy) = - \log q(\vy \mid \vf)$.
\Eg, for square loss, the net's likelihood is a Gaussian, and for softmax cross-entropy loss a categorical distribution.
This allows the introduction of an alternative notion of distance based on the KL divergence between the likelihoods implied by two models $f_{\vtheta}$ and $f_{\vtheta + \vdelta}$.
The metric of this statistical manifold is the Fisher~\cite{amari2000natural}.
Following \citet{soen2024tradeoffs}, we consider two common forms of the Fisher. Here we describe the type-I Fisher which we simply denote as Fisher. The overview of type-II Fisher is in \cref{sec:curvature_matrices_more}.

\paragraph{Fisher.} The Fisher matrix is the log likelihood's gradient covariance \wrt the model's distribution $q$.
Denoting the log-likelihood as $l_n \coloneqq \log q(\vy \mid \vf_n)$, we have
\begin{align}
\label{eq:fisher-type-I}
   &\textstyle\mF_{\text{I}}(\vtheta)
  \coloneqq
  R_{\sD} \;
  \sum_{\vx_n \in \sD}
  \E_{q(\vy \mid \vf_n)}
  \left[
    \nabla_{\vtheta} l_n
    \left(
    \nabla_{\vtheta} l_n
    \right)^{\top}
    \right]
  \\ \nonumber
   &\textstyle \phantom{:}=
  R_{\sD} \;
  \sum_{\vx_n \in \sD}
  {\mathrm{J}_{\vtheta} \vf_n}^{\top}
  \E_{q(\vy \mid \vf_n)}
  \left[
    \nabla_{\vf_n} l_n
    \left(
    \nabla_{\vf_n} l_n
    \right)^{\top}
    \right]
  \mathrm{J}_{\vtheta} \vf_n\,.
\end{align}
\paragraph{Monte-Carlo Fisher.} In practice, we approximate the expectation in \eqref{eq:fisher-type-I} through Monte-Carlo (MC) sampling. 
To this end, let $\smash{\hat{\vy}_{n,s} \stackrel{\text{i.i.d.}}{\sim} q(\vy \mid \vf_n)}$ and denote $\hat{l}_{n,s} \coloneqq \log q(\vy = \hat{\vy}_{n,s} \mid \vf_n)$ the would-be log-likelihood on that sample. 
Then, the MC-estimated Fisher with $S$ samples is
\begin{align}
  \label{eq:fisher-mc}
  \begin{split}
    \hat{\mF}_{\text{I}}(\vtheta)
     & =
    \textstyle\nicefrac{R_{\sD}}{S}
    \sum_{\vx_n \in \sD}
    \sum_{s=1}^{S}
    \nabla_{\vtheta} \hat{l}_{n,s}
    \left(
    \nabla_{\vtheta} \hat{l}_{n,s}
    \right)^{\top}\,.
  \end{split}
\end{align}

\paragraph{Empirical Fisher (gradient covariance).} 
Another popular modification of~\eqref{eq:fisher-type-I} is to replace the expectation over the model's likelihood $q(\vy \mid \vf_n)$ with the empirical likelihood implied by the data, $q_{\text{emp}}(\vy \mid \vf_n) = \delta(\vy - \vy_n)$ where $\vy_n$ is the true label. This yields the empirical Fisher,
\begin{align}\label{eq:empirical-fisher}
  \begin{split}
  \hspace{-5pt}
    \mE(\vtheta)
     & \textstyle\coloneq
    R_{\sD} \;
    \sum_{\vx_n \in \sD}
    \E_{q_{\text{emp}}(\vy \mid \vf_n)}
    \left[
      \nabla_{\vtheta} l_n
      \left(
      \nabla_{\vtheta} l_n
      \right)^{\top}
      \right]
    \\
     & \textstyle\phantom{:}=
    R_{\sD} \;
    \sum_{(\vx_n, \vy_n) \in \sD}
    \nabla_{\vtheta} \ell(\vf_n, \vy_n)
    \left(
    \nabla_{\vtheta} \ell(\vf_n, \vy_n)
    \right)^{\top}\,,
  \end{split}
\end{align}
the (scaled) un-centered gradient covariance \wrt the empirical data distribution.
The empirical Fisher differs from the Fisher~\cite{kunstner2019limitations}, but is popular in applications as it re-uses information from gradient backpropagation.

\paragraph{Fisher-vector products (FVPs) via GGNVPs.}
The na\"ive approach to multiply with the Fisher and empirical Fisher is to compute the (would-be) per-datum gradients.
However, they require $|\sD| \times D$ additional memory.
Instead,
\texttt{curvlinops} utilizes the similarity of these matrices to the GGN: 
FVPs are done via GGNVPs on a pseudo-loss, which is more efficient.
To see this, note that the last lines of \eqref{eq:fisher-mc} \& \eqref{eq:empirical-fisher} resemble the GGN in \eqref{eq:ggn}.
Construct the pseudo-loss $\tilde{\ell}(\vf_n, \vy_n) = \vf_n^{\top}\vg_n\vg_n^{\top}\vf_n$ where $\vg_n$ is a detached vector whose value is set to $\nabla_{\vf_n} \log q(\hat{\vy}_{n,s} \mid \vf_n)$ or $\nabla_{\vf_n} \ell(\vf_n, \vy_n)$.
The GGN of that pseudo-loss then corresponds to the desired Fisher, as $\smash{\nabla_{\vf_n}^2 \tilde{\ell}(\vf_n, \vy_n) = \vg_n \vg_n^{\top}}$ in \eqref{eq:ggn}.
The linear operator interface used in \texttt{curvlinops} abstracts away these implementation details, allowing the user to focus on higher-level tasks without worrying about the specifics.

\paragraph{Kronecker-Factored Approximate Curvature (KFAC).} 
KFAC~\cite{heskes2000natural,martens2015optimizing} is a light-weight parametric curvature proxy that uses Kronecker products to approximate per-layer curvature matrices, \ie let $\vtheta^{(l)}$ be the parameters residing in layer $l$ of a network, then KFAC approximates the curvature matrix $\mC(\vtheta^{(l)}) \approx \mA^{(l)} \otimes \mB^{(l)}$ where $\mA^{(l)}$ is computed from the layer's input, and $\mB^{(l)}$ from (would-be) gradients \wrt to the layer's output. 
KFAC applies to the GGN \cite{botev2017practical}, MC, and empirical Fisher and a broad range of neural nets~\cite{grosse2016kroneckerfactored,martens2018kroneckerfactored,eschenhagen2023kroneckerfactored}.
It was further improved through eigenvalue correction \citep[EKFAC,][]{george2018fast}.
Implementing these variations and ensuring their correctness is highly complex.

\texttt{curvlinops} provides linear operators for the Hessian, GGN, MC Fisher, empirical Fisher, and various (E)KFAC flavors.
We also provide the damped inverses of (E)KFAC, with support for the empirical damping heuristic of \citet{martens2015optimizing}, and the exact damping scheme used by \citet{grosse2023studying}.
\texttt{curvlinops} also supports backpropagation-free input-based curvature from \citet{benzing2022gradient,petersen2023isaac}, and the KFAC-reduce approximation from~\citet{eschenhagen2023kroneckerfactored} for neural networks with linear weight sharing layers.

\subsection{Convenience Functionality \& Safeguards}\label{subsec:convenience-and-safeguards}

So far, we have seen the complexity introduced by automatic differentiation, probabilistic interpretation of loss functions and MC sampling, as well as the details of computing KFAC variations.
Here, we present additional conveniences that can be incorporated into a linear operator interface.

\paragraph{Automatically handling batch aggregation.}
For computations, it is common to batch empirical risks (\Cref{eq:empirical-risk}) by splitting the data into $N$ disjoint batches, $\sD = \sB_1 \cup \sB_2 \cup \ldots \cup \sB_N$, and evaluate the empirical risk on $\sD$ by accumulation over the batch losses $\gL_{\sB_i}(\vtheta)$,
\begin{align}\label{eq:empirical-risk-batches}
  \textstyle\gL_{\sD}(\vtheta)
  =
  \sum_{i=1}^N
  \nicefrac{R_{\sD}}{R_{\sB_i}} \gL_{\sB_i}(\vtheta)\,,
\end{align}
with $\gL_{\sB_i}(\vtheta)$ defined via \Cref{eq:empirical-risk} for the batch $\sB_i$.
The aggregation over batches requires correcting the batch scaling and accounting for the global scaling.

Curvature computations can be split into batches in exactly the same way.
Automatically accounting for correct aggregation over batches is useful, because it allows to change the batch size or data split without changing the matrix-vector product.

\paragraph{Automatically handling parameter formats.}
We often regard the neural net parameters $\vtheta \in \sR^D$ as a vector.
However, in software, they are separated into tensors in different layers and roles, such as weights and biases.
\Eg, for a multi-layer perceptron $f_\vtheta(\vx) \coloneqq \vx^{(L)}$ with $L$ layers and element-wise activation function $\sigma$, where $\vz^{(l)} = \sigma \left( \mW^{(l)} \vz^{(l-1)} + \vb^{(l)} \right)$ for $l = 1, \dots, L$  and $\vz^{(0)} = \vx$, the weights $\mW^{(l)}$ are matrices and the biases $\vb^{(l)}$ are vectors.
The total parameter space is a tensor product of tensor lists $(\mW^{(1)}, \vb^{(1)}, \dots, \mW^{(L)}, \vb^{(L)}) \simeq \vtheta$.
To obtain the vector representation, one simply flattens and concatenates the entries, $\vtheta = \Cat(\flatten \mW^{(1)}, \vb^{(1)}, \dots, \flatten \mW^{(L)}, \vb^{(L)})$.

A linear operator $\texttt{A}$ for a curvature matrix can process both formats and return the result in the same format:
\begin{lstlisting}[%
  language=python,%
  xleftmargin=0ex,%
  numbers=none,%
]
 # 1) input and result as vector
 v = torch.rand(A.shape[1])
 Av = A @ v
 # 2) input and result as tensor list
 v = [torch.rand_like(p) for p in params]
 Av = A @ v
\end{lstlisting}%

\paragraph{Preventing common mistakes.}
Many deep learning techniques introduce stochasticity into the neural net, and therefore the empirical risk~\eqref{eq:empirical-risk}.
Others induce non-deterministic, or data-order dependent behavior, in the batched empirical risk~\eqref{eq:empirical-risk-batches}.
This is undesirable: We want a linear operator $\gA$ to \emph{always} return the same $\gA \vv$ for a fixed $\vv$, independent of the batching scheme or the random number generator state.
A non-deterministic linear operator introduces numerical inaccuracies into our application, e.g.\,CG becomes unstable with a noisy matrix~\cite{martens2010deep}.
Examples of techniques that can introduce non-deterministic behavior are data augmentation, dropout~\cite{srivastava2014dropout}, and batch normalization~\cite{ioffe2015batch}.
Although such pathologies are hard to detect in general, \texttt{curvlinops} provides a sanity check that identifies most failure scenarios and significantly reduces the risk of introducing such spurious artifacts:
When creating a linear operator, it (optionally) checks whether two consecutive evaluations of the empirical risk, its gradient, and the operator's matrix-vector product are identical.

\paragraph{Ensuring correctness.}
Testing some curvature approximations, specifically, KFAC, is challenging because they only become exact in special settings.
Without testing, this can easily introduce scaling bugs where a Kronecker factor is off by a scalar.
Our KFAC implementation tests KFAC's equivalence to the block-diagonal GGN, empirical Fisher, or type-I Fisher using the scenarios described in~\cite{bernacchia2018exact,eschenhagen2023kroneckerfactored}.
Although these scenarios are not comprehensive, this is an active field of research, and further tests can be added to \texttt{curvlinops} as new equivalences and approximations are discovered.

\begin{comment}
\textbf{Block-diagonal approximations.} \quad
Many curvature approximations ignore dependencies between parameters in different layers, which corresponds to the diagonal blocks of the associated curvature matrix. 
Moreover, there has recently been some interest in the properties of Hessian block diagonals and their possible implications for optimization \citep{zhang2024transformers,zhang2024adam}.
To simplify the relation of such block-diagonal proxies to their exact counterparts (as e.g.\,done by \citet{wu2020dissecting,ormaniec2024transformer}), or other applications like block-diagonal Hessian-free optimization \cite{zhang2017blockdiagonal}, some linear operators in \texttt{curvlinops} support block-diagonal approximations; for example the Hessian (see \Cref{fig:visual-tour}).
Parameters can freely be assigned to a block.
%
%
%
\end{comment}

%
%
%
%

\section{Linear Operators Simplify Applications of Curvature Matrices}\label{sec:transferability}
Here, we argue that linear operators simplify using curvature matrices in a broad range of applications.
The workflow simplifies into three steps.

\textbf{Step 1: Create a linear operator.} \quad
Recall from \Cref{sec:encapsulation} that an empirical risk, or one of its associated curvature matrices, requires four ingredients: a neural net architecture, the loss function, a data set, and the network parameters.
A unified interface for all curvature matrices addressed in \Cref{sec:encapsulation} is:
\begin{lstlisting}[%
  language=python,%
  xleftmargin=0ex,%
  numbers=none,%
]
 A = LinearOperator(
   model, # Neural net, $f_{\vtheta}: \{\vx_n\}_{\sB} \mapsto \{ \vf_n \}_{\sB}$
   loss_func, # Criterion $\ell$, $(\{\vf_n\}_{\sB}, \{\vy_n\}_{\sB}) \mapsto \gL_{\sB}(\vtheta)$
   params, # $\vtheta$ as list, $\smash{(\mW^{(1)}, \vb^{(1)}, \dots, \mW^{(L)}, \vb^{(L})}$
   data, # Data loader, yields a batch $(\{\vx_n\}_{\sB}, \{\vy_n\}_{\sB})$
   check_deterministic=True, # Safeguards $\text{\Cref{subsec:convenience-and-safeguards}}$
   **kwargs, # Optional configuration arguments
 )
\end{lstlisting}%

\textbf{Step 2: Compose or transform operators.} \quad
Linear operators can lazily be composed, \eg summed, or transformed to represent or approximate other matrix functions of $\gA$ (\eg via Lanczos/Arnoldi).
One specific example is inverting a linear operator, \ie creating a linear operator representing $\gA^{-1}$ given $\gA$.
Note that the product $\vv \mapsto \gA^{-1} \vv \coloneqq \vz$ requires solving $\gA \vz\ = \vv$ for $\vz$. This can be done with an iterative solver based on matrix-vector products with $\gA$, like truncated CG: $\gA^{-1} \vv = \texttt{cgsolve(}\gA, \vv\texttt{)}$. Alternatives are using a truncated Neumann series $\gA^{-1} = \sum_{k=0}^{\infty} \left( \mI - \gA \right)^k$ or, in the case of KFAC, leveraging properties of the Kronecker product structure.
Since transformed or composed operators are again linear operators, they can be used in exactly the same way as the operators we started out with.

\textbf{Step 3: Use operators like dense matrices.} \quad
We will discuss this next along with a series of relevant downstream applications to showcase the transferability of linear operators and provide associated code snippets that demonstrate the similarity between code and mathematical notation.

\subsection{Examples}

\textbf{Application 1: Second-order optimization.} \quad
Newton's method and natural gradient descent (NGD) precondition a gradient vector $\vg(\vtheta)$ with the inverse of a curvature matrix $\mC(\vtheta)$: $\vtheta \leftarrow \vtheta - \eta \mC(\vtheta)^{-1} \vg(\vtheta)$.
Due to the size of $\mC(\vtheta)$, inversion can only be done implicitly, or through structural approximations that are cheap to store or invert.

Here are three different approaches for computing approximate Newton or natural gradient steps:
\begin{lstlisting}[%
basicstyle={\small\ttfamily},
language=python,%
xleftmargin=0ex,%
numbers=none,%
]
g = gradient()
# Newton-CG $\text{\rmfamily\cite{martens2010deep}}$
G = GGNLinearOperator(...)
G_inv = CGInverseLinearOperator(G)
step = -G_inv @ g
# Neumann optimizer $\text{\rmfamily\cite{krishnan2017neumann}}$
H = HessianLinearOperator(...)
H_inv = NeumannInverseLinearOperator(H)
step = -H_inv @ g
# KFAC optimizer $\text{\rmfamily\cite{martens2015optimizing}}$
K = KFACLinearOperator(...)
K_inv = KFACInverseLinearOperator(K)
step = -K_inv @ g
\end{lstlisting}%

\textbf{Application 2: Influence functions.}\quad 
Inverse Hessian-vector products (IHVPs), or more generally inverse curvature-vector products also occur during differentiation through an empirical risk minimization procedure.
One important application is influence functions~\cite{hampel1974influence,koh2017understanding,grosse2023studying,mlodozeniec2025influence}, which study a perturbation's impact on the minimizer
\vspace{-3pt}
\begin{equation*}
  \textstyle\hat{\vtheta}(\vtheta, \epsilon)
  =
  \argmin_{\vtheta}
  \left(
  \gL_{\sD}(\vtheta) + \epsilon P(\vtheta)
  \right)
\end {equation*}
where $P(\vtheta)$ is a perturbation (\eg up/down-weighting, or perturbing, a data point).
Then, the implicit function theorem provides the differentiation rule for $\hat{\vtheta}$ without having to differentiate through the unrolled optimization procedure. The perturbation's influence on the parameters is \( \gI = \left.\nicefrac{\mathrm{d}\hat{\vtheta}}{\mathrm{d}\epsilon}\right|_{\epsilon=0} = \mH(\hat{\vtheta})^{-1}  \nabla_\vtheta P(\hat{\vtheta})\). \Eg, the influence for up-weighting data point $n$ ($P(\vtheta) = R_\sD \ell(f_\vtheta(\vx_n), \vy_n)$) is
\begin{lstlisting}[%
basicstyle={\small\ttfamily},
language=python,%
xleftmargin=0ex,%
numbers=none,%
]
g_n = gradient(...)
# influence using EKFAC $\text{\rmfamily\cite{grosse2023studying}}$
EK = EKFACLinearOperator(...)
EK_inv = EKFACInverseLinearOperator(EK)
I = -EK_inv @ g_n
\end{lstlisting}%

Similar hyper-gradients also occur in algorithms for bi-level \cite{lu2022indiscriminate,evtushenko1974iterative} and hyper-parameter optimization \cite{lorraine2020optimizing}.

\textbf{Application 3: Model merging.} \quad 
Assume we are given a neural network architecture $f_\vtheta$ that is trained on $T$ independent tasks $\sD_1, \dots, \sD_T$ using loss functions $\ell_1, \dots, \ell_T$. This yields $T$ parameter vectors $\vtheta_1, \dots, \vtheta_T$ which we are supposed to merge into a single vector $\vtheta_\star$.
One na\"ive way is to simply average the parameters, $\vtheta_\star = \nicefrac{1}{T} \sum_{t=1}^T \vtheta_t$.
Fisher-weighted model merging \cite{matena2022merging} is a superior, albeit more expensive, merging procedure where each parameter $\vtheta_t$ is weighted by the Fisher matrix $\mF_t(\vtheta_t)$ evaluated on the task $\sD_t$, then normalized:
\vspace{-4pt}
\begin{align*}
    \textstyle\vtheta_\star
    =
    \left(
    \sum_{t=1}^T \mF_t(\vtheta_t)
    \right)^{-1}
    \left(
    \sum_{t=1}^T
    \mF_t(\vtheta_t) \vtheta_t
    \right)\,.
\end{align*}
For computational reasons, the original work resorts to a diagonal approximation of the Fisher. 
Here is how to implement the original approach, and an alternative unexplored variant that uses the full instead of the diagonal Fisher:
\begin{lstlisting}[%
basicstyle={\small\ttfamily},
language=python,%
xleftmargin=0ex,%
numbers=none,%
]
params, ls, Ds = [...], [...], [...]
# Per-task Fisher matrices
Fs = [GGNLinearOperator(..., l_t, p_t, D_t)
      for p_t, l_t, D_t
      in zip(params, ls, Ds)]
# Diagonal Fisher merging $\text{\rmfamily\cite{matena2022merging}}$
diag_Fs = [hutchinson_diag(F_t)
           for F_t in Fs]
sum_inv = sum(diag_Fs) ** (-1)
merge = sum_inv * sum(F_t * p_t
    for F_t, p_t in zip(diag_Fs, params))
# Full Fisher merging via CG (unexplored)
sum_inv = CGInverseLinearOperator(sum(Fs))
merge = sum_F_inv @ sum(F_t @ p_t
    for F_t, p_t in zip(Fs, params))
\end{lstlisting}%

\textbf{Application 4: Model pruning/sparsification.}
With the advent of larger parameter spaces, interest in pruning arose.
One idea is to eliminate parameters that least impact the loss~\cite{lecun1889optimal,hassibi1992second}.
Given a neural net trained to convergence at $\vtheta$, the optimal perturbation $\tilde{\vtheta} = \vtheta+\vdelta(i)$, such that $[\tilde{\vtheta}]_i\!=\!0$ with minimal increase ($\rho$) of the loss is
\begin{align*}
\vdelta (i) = \frac{-[\vtheta]_i \mH(\vtheta)^{-1} \ve_i}{[\mH(\vtheta)^{-1}]_{i,i}}, \quad \rho(\vdelta(i)) = \frac{[\vtheta]_i^2}{2 [\mH(\vtheta)^{-1}]_{i,i}}\,.
\end{align*}
This procedure is costly, requiring IHVPs and the inverse Hessian's diagonal. 
Recent work resorted to KFAC to scale similar ideas to large language models \cite{ouderaa2024llm}. Another popular approximation, showcased below, is to assume diagonality, \ie $[\mH^{-1}(\vtheta)]_{i,i}\!=\!\nicefrac{1}{\mH(\vtheta)_{i,i}}$:
\begin{lstlisting}[%
basicstyle={\small\ttfamily},
language=python,%
xleftmargin=0ex,%
numbers=none,%
]
# OBS pruning with approximate Hessian diagonal
H = HessianLinearOperator(...)
diag_H_inv = 1 / hutchinson_diag(H)
rho = params**2 / (2 * diag_H_inv)
# prune 5% of parameters
set_zero = int(0.05 * params.numel())
_, idx = rho.topk(set_zero, largest=False)
params[idx] = 0.0
\end{lstlisting}%

\textbf{Application 5: Loss landscape analysis.}\quad
Many works have studied the spectrum of curvature matrices~\cite{ghojogh2019eigenvalue,yao2020pyhessian,sagun2017eigenvalues,sagun2018empirical,papyan2019measurements}.
As an example, we consider the phenomenon observed by \citet{gurari2018gradient} that the gradient resides in the Hessian's top eigenspace.
This is measured by overlap, which projects the gradient $\vg(\vtheta)$ onto the space spanned by the top eigenvectors, yielding $\vg(\vtheta)_\text{proj}$, then forming $\nicefrac{\lVert \vg(\vtheta)_\text{proj} \rVert^2}{ \lVert \vg(\vtheta) \rVert^2} \in [0; 1]$.
This can be done as follows:
\begin{lstlisting}[%
basicstyle={\small\ttfamily},
language=python,%
xleftmargin=0ex,%
numbers=none,%
]
g = gradient().numpy()
H = HessianLinearOperator(...).to_scipy()
# Gradient overlap with top eigenspace $\text{\rmfamily\cite{gurari2018gradient}}$
_, evecs = eigsh(H, k=10)
g_top = sum(dot(e, g) * e for e in evecs.T)
overlap = norm(g_top) ** 2 / norm(g) ** 2
\end{lstlisting}%
In the above snippet, we have already used some interoperability with SciPy to leverage its highly sophisticated sparse eigensolvers.
This capability, which we elaborate on in the next section, completes our argument in favor of using linear operators to compute with curvature matrices.

\textbf{Remark.} \quad
The list above is non-exhaustive and any applications that use the Hessian or approximations thereof can enjoy the benefits of the argued linear operator abstraction.
One prominent example is uncertainty quantification with Bayesian neural nets.
\Eg, \citet{yang2024laplacelora,kristiadi2024sober} show that the Kronecker-factored GGN is useful for turning standard large language models into Bayesian ones, enabling downstream applications in Bayesian optimization.

\section{Linear Operators Enable Extensibility \& Interoperability}\label{sec:interoperability}
For the last part of our argument, we emphasize the potential of linear operators to foster deep learning research by facilitating the connection between curvature matrices and the field of randomized linear algebra (RLA), as well as by enabling the application of sophisticated linear operator routines from libraries like SciPy~\cite{virtanen2020scipy}.

\paragraph{Bridge to randomized linear algebra.}
RLA is a mature field, leveraging sharp inequality bounds that lead to algorithms with remarkable stability and scalability, spanning numerous applications like dimensionality reduction, combinatorial optimization, compressed sensing, and more \citep{tropp2015introduction,brailovskaya2024inequalities}.
Despite its relevance, the adoption of RLA into ML research is behind other computational fields.
This issue was recently highlighted in a 2023 NeurIPS tutorial, pointing at challenges involving hardware, software and theory-practice gaps \cite{derezinski2024recent}.
We think linear operators help overcome many of these practical issues due to their simple and universal interface that is \emph{compatible} with such algorithms:
ML practitioners wishing to incorporate an existing RLA procedure just have to worry about satisfying this interface from both sides, with minimal overhead.
To demonstrate this last point, we implemented some methods from the literature that are compatible with any linear operator in \texttt{curvlinops} (see \cref{fig:matrix-properties} for some toy experiments):
\begin{itemize}
  \item \textbf{Spectral densities (\Cref{fig:matrix-properties}, top).} Let $\lambda_i$ denote the $i$th eigenvalue of $\gA$.
  The spectral density of $\gA$ is $\rho(\lambda) = \nicefrac{1}{\dim(\gA)} \sum_i \delta(\lambda - \lambda_i)$.
    We implement two algorithms from \citet{papyan2020prevalence} that are based on Lanczos iterations.
        One estimates $\rho(\lambda)$, the other estimates the matrix logarithm's spectral density
        $\rho_{\text{log}}(\nu) = \nicefrac{1}{\dim(\gA)} \sum_i \delta(\nu - \log(|\lambda_i| + \epsilon))$
        with $\epsilon > 0$, which is the spectral density of $\log\left(|\gA| + \epsilon \mI\right)$ where $| \gA |$ is the absolute value matrix function.

  \item \textbf{Trace, squared Frobenius norm (\Cref{fig:matrix-properties}, bottom left).} The simplest method to estimate the trace $\Tr(\gA)$ is the Girard-Hutchinson method \cite{girard1989montecarlo,hutchinson1989stochastic}, which draws \iid vectors from a distribution with zero mean and unit covariance,
  and uses that $\Tr(\gA) = \Tr(\gA \E[\vv \vv^\top]) = \E[\vv^\top \gA \vv]$.
  We also implement its variance-reduced version Hutch++ \cite{meyer2020hutch}, and XTrace \cite{epperly2024xtrace}, which combines variance reduction with the exchangeability principle to achieve invariant estimates when permuting test vectors.
  All are unbiased, and variance reduction yields more accurate trace estimators for matrices with fast spectral decay whose trace is dominated by a few eigenvalues, at the cost of additional memory.

  Trace estimation allows estimating the squared Frobenius norm $\lVert \gA \rVert_{\text{F}}^2 = \Tr(\gA \gA^{\top})$ as well.
  We also implement a variant that uses half the matrix-vector products.

  \item \textbf{Diagonal (\Cref{fig:matrix-properties}, bottom right).} \citet{bekas2007estimator} extended the trace estimation idea from Hutchinson's method to matrix diagonals $\diag(\gA)$.
        We implement their algorithm and XDiag from \citet{epperly2024xtrace}, which generalizes XTrace to diagonals.
        Similar to the trend for trace estimation, we observe XDiag to be more accurate on matrices with fast spectral decay.
\end{itemize}
We have included some of these methods in the code snippets in \Cref{sec:transferability}, illustrating how linear operators provide a bridge to RLA, making RLA accessible to a broader audience.

\paragraph{SciPy/NumPy compatibility.}
Many other RLA routines are already implemented in other established packages, like SciPy/NumPy~\cite{harris2020array,virtanen2020scipy}. To avoid their re-implementation, \texttt{curvlinops}'s linear operators can easily be exported to SciPy linear operators that consume NumPy arrays:%
\begin{lstlisting}[%
  language=python,%
  xleftmargin=0ex,%
  numbers=none,%
]
 # 3) SciPy/NumPy support
 A = A.to_scipy()
 v = numpy.random.rand(A.shape[1])
 Av = A @ v
\end{lstlisting}%
This allows executing the expensive curvature matrix-vector multiplies on a GPU, while unlocking SciPy's highly sophisticated routines for linear operators, many of which currently lack a compatible interface in PyTorch.
One example is \texttt{scipy.sparse.linalg}'s highly efficient iterative solvers for partial eigen- and singular value decompositions like \href{https://docs.scipy.org/doc/scipy/reference/generated/scipy.sparse.linalg.eigsh.html}{\texttt{eigsh}} and \href{https://docs.scipy.org/doc/scipy/reference/generated/scipy.sparse.linalg.svds.html}{\texttt{svds}} based on implicitly restarted Arnoldi iterations~\cite{lehoucq1998arpack}.
These sophisticated solvers usually require much fewer matrix-vector products compared to na\"ive methods, like the power iteration, and are not (yet) implemented in PyTorch or JAX.
Other examples are linear system solvers such as conjugate gradients \citep[CG,][]{hestenes1952methods} or LSMR~\cite{fong2010lsmr}.

\paragraph{Bridge to under-explored sketching algorithms for DL.}
In addition to the previously mentioned methods, there is a growing number of cutting-edge RLA algorithms that find their way into deep learning thanks to linear operators.
For example, \citet{singh2021analytic,singhal2023guess} show that curvature matrices like the Hessian can be highly rank-deficient, making low-rank decompositions valuable to approximate curvature.
Still, computing $k$ such eigenvectors using an iterative solver requires $\bigO(\tau k)$ HVPs, across $\tau$ \emph{sequential} iterations \citep[][7.3]{golub2013matrix} becomes intractable for larger nets;
 numerical stability can be an issue, too.
In contrast, sketch-and-solve methods require only $\bigO(k)$ \emph{parallelizable} HVPs, and are numerically stable, allowing to scale low-rank approximations \citep{halko2011svd, tropp2019streaming}.
The \texttt{skerch} library \citep{fernandez2024skerch} implements
such sketched matrix decompositions, assuming a minimal linear operator interface in PyTorch.
This allows to
plug-in \texttt{curvlinops} with minimal overhead:
\begin{lstlisting}[%
  language=python,%
  xleftmargin=0ex,%
  numbers=none,%
]
H = HessianLinearOperator(...)
# Low-rank sketch: $\smash{\mH \approx \mQ \mU \mS \mU^\top \mQ^\top}$
Q, U, S = skerch.seigh(H, ...)
\end{lstlisting}%

\paragraph{Bridge to optimizers \& compositional linear algebra.}
Many (inverse) curvature approximations are developed in the context of second-order optimization and are estimated during training.
Linear operators can potentially export these approximations at any point of training, to leverage them for any of the applications we mention in \cref{sec:transferability}.
Realizing this idea requires the optimizer to implement the (inverse) curvature matrix as a linear operator.
We demonstrate this approach and implement linear operators for the structured and inverse-free KFAC~\cite{lin2024structured} and inverse-free Shampoo~\cite{lin2024can} optimizers.
These optimizers offer linear operators of block-diagonal Kronecker-factored estimators of the empirical Fisher and gradient outer product inverses.
Other algorithms whose curvature approximation could be made available via linear operators in the future are improved versions of the empirical Fisher for fine-tuning~\cite{wu2024improved}, Ginger~\cite{hao2024ginger}, M-FAC~\cite{frantar2021mfac}, or Spectral~\cite{lin2024fast}.
Finally, to leverage the latest structure-aware linear operator implementations via dispatch algorithms, one could consider supporting interoperability with the CoLA package~\cite{potapczynski2023cola}.

\section{Alternative Views}

Some benefits of presenting curvature matrices as linear operators can also be seen as potential pitfalls. %

\paragraph{Encapsulation may lead to duplicated computation.}
Encapsulating all computations related to the curvature matrix in a linear operator (\cref{sec:encapsulation}) means in practice that it might be necessary to duplicate computations.
For example, in the case of a gradient descent algorithm that preconditions the gradient with a KFAC approximation of the empirical Fisher, we theoretically only require a single forward-backward pass to compute the gradient and the preconditioner.
However, if the preconditioner computation is isolated in a linear operator, it might be necessary to compute an additional redundant forward-backward pass, which doubles the cost.

\paragraph{Hiding complexity may promote incorrect usage.} While hiding the complexity of curvature matrices makes it easier to apply them to new applications (\cref{sec:transferability}), this might also promote their incorrect usage.
How exactly a curvature matrix should be defined for a specific application is non-trivial, e.g., the GGN is determined by choosing a specific \emph{split} of the objective \cite{martens2020new,kunstner2019limitations}.
When exposing a curvature matrix as a linear operator to users, the implementation will have to be on a spectrum between two extremes to handle this nuance:
(1) the interface can include configuration options to cover as many variations of the curvature matrix as possible, and (2) only a single variation of the curvature matrix is supported, based on what is deemed the most broadly applicable set of assumptions.
Case (1) requires expert knowledge of the user, although admittedly the implementation burden is still lifted from them.
In case (2) the curvature matrix required by the application might not even be available.
In both cases, the user might unknowingly use a curvature linear operator that is inappropriate for their application.

\paragraph{Linear operator interface may be too minimal.}
Finally, while many applications of curvature matrices only require matrix-vector products, others might benefit from the ability to directly access sub-matrices of a curvature matrix.

\section{Conclusion}
We have advocated for the use of linear operators as a powerful abstraction for handling structured matrices, particularly curvature matrices, in machine learning applications.
Presenting these matrices as linear operators enables users to leverage their benefits across various applications without the burden of implementation complexity.
This approach not only encapsulates the intricacies of curvature computations but also enhances extensibility and interoperability, as demonstrated through our \texttt{curvlinops} library.
Our work underscores the potential of linear operators to democratize access to advanced structural approximations, facilitating their integration into large-scale neural network architectures and transferring techniques to other machine learning tasks.
We believe that embracing curvature matrices represented as linear operators will be crucial for advancing both research and practical applications in deep learning.

\begin{comment}
\paragraph{Future work.}
One important limitation which we plan to address in the near future is adding multi-GPU support.
We also plan to add new curvature approximations, for instance the GGN's hierarchical decomposition from~\citet{papyan2020prevalence} and the Shampoo preconditioner~\cite{gupta2018shampoo}.
It might also be interesting to migrate to inheriting the linear operator base class from that defined in \citet{potapczynski2023cola} or \citet{gardner2018gpytorch} to support compositional rules out of the box.
Another, rather exotic feature would be to make the linear operators fully-differentiable.

Long-term, it might be interesting to consider migrating to this interface, as it implement various compositional rules, and structural linear algebra tricks.

Differentiable matrix functions via implicit differentiation through Lanczos/Arnoldi~\cite{kraemer2024gradients}.

Maybe: think more about how it can be used for/within optimizers (maybe just out of scope).

%
%
%
%
%
\end{comment}

%
%
%
%

%
\ifdefined\isaccepted
\section*{Acknowledgements}
The authors would like to thank Jonathan Wenger for providing feedback to the manuscript.
Resources used in preparing this research were provided, in part, by the Province of Ontario, the Government of Canada through CIFAR, and companies sponsoring the Vector Institute.
Runa Eschenhagen is supported by ARM, the Cambridge Trust, and the Qualcomm Innovation Fellowship.
Andres Fernandez is supported by the DFG through Project HE 7114/5-1 in SPP2298/1.
The authors thank the International Max Planck Research School for Intelligent Systems (IMPRS-IS) for supporting Andres Fernandez and Lukas Tatzel, and gratefully acknowledge co-funding by the European Union (ERC, ANUBIS, 101123955).
Views and opinions expressed are however those of the author(s) only and do not necessarily reflect those of the European Union or the European Research Council.
Neither the European Union nor the granting authority can be held responsible for them.
We also gratefully acknowledge the German Federal Ministry of Education and Research (BMBF) through the Tübingen AI Center (FKZ: 01IS18039A); and funds from the Ministry of Science, Research and Arts of the State of Baden-Württemberg
\fi

\bibliography{references.bib}
\bibliographystyle{icml2025}

\clearpage
\appendix

\section{Performance}\label{sec:performance}
\begin{figure*}[!ht]
  \centering
  \begin{minipage}{0.01\linewidth}
    \rotatebox{90}{\textbf{ResNet50 on ImageNet}}
  \end{minipage}
  \begin{minipage}{0.49\linewidth}
    \centering
    \textbf{Run time}\vspace{1ex}
    \includegraphics{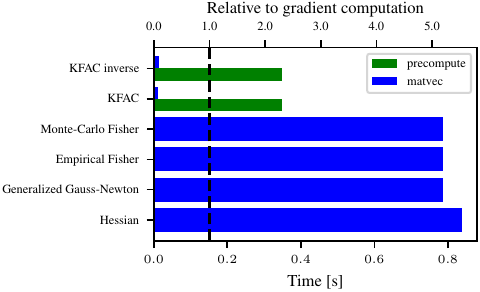}
  \end{minipage}
  \hspace{-1.5ex}
  \begin{minipage}{0.49\linewidth}
    \textbf{Peak memory}\vspace{1ex}
    \centering
    \includegraphics{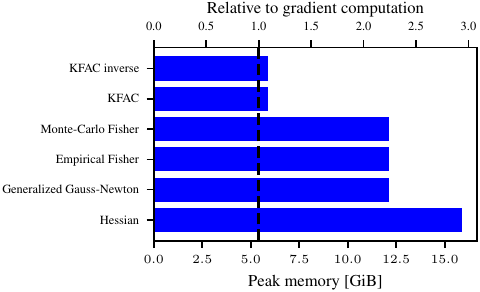}
  \end{minipage}

  \vspace{1ex}

  \begin{minipage}{0.01\linewidth}
    \rotatebox{90}{\textbf{nanoGPT on Shakespeare}}
  \end{minipage}
  \begin{minipage}{0.49\linewidth}
    \centering
    \includegraphics{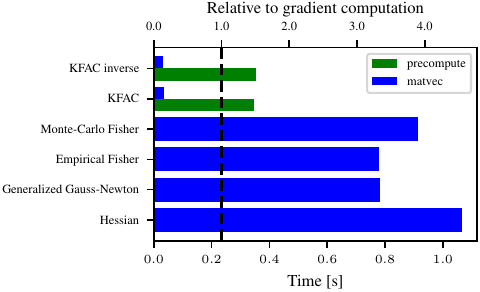}
  \end{minipage}
  \hspace{-1.5ex}
  \begin{minipage}{0.49\linewidth}
    \centering
    \includegraphics{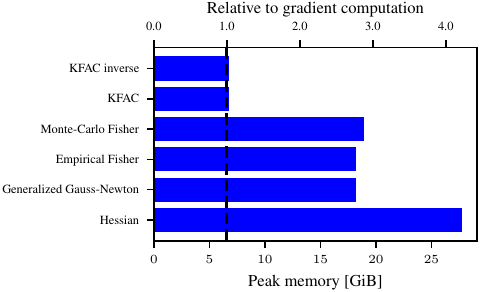}
  \end{minipage}

  \caption{\textbf{Performance analysis:} Run time (left column) and peak memory (right column) of linear operators benchmarked on ResNet50 on ImageNet (top row) and nanoGPT on Shakespeare (bottom row) on an A40 GPU with 40\,GiB of RAM (the code used to generate these results is \href{\linkToGithub/blob/75bc0a84b2001f052daeaeac9a58846f379fed8a/docs/examples/basic_usage/example_benchmark.py}{here}).
    \textbf{Details:} For ImageNet, we use a batch size of $64$ and images of shape $(3, 224, 224)$; for Shakespeare, we use a batch size of $4$ and context length $1024$.
    All linear operators use their default options and we do not use compilation. Models are in evaluation mode.
    KFAC neglects parameters in normalization layers (they are unsupported), and nanoGPT's last layer due to its large dimension ($\approx$50\,K).
  }
  \label{fig:performance}
\end{figure*}

Here, we evaluate the performance of \texttt{curvlinops}'s linear operators in terms of run time and memory consumption.
This serves three purposes:
(1) the results can be used as a rule of thumb by users to estimate the computational cost relative in terms of multiples of the gradient;
(2) it demonstrates that \texttt{curvlinops} indeed scales to large models,
and (3) it allows to highlight some PyTorch-specific limitations which, if addressed, enable future improvements.

\paragraph{Setup} We consider two representative tasks for large-scale image recognition and language modelling: ResNet50 \citep[$\approx$26\,M parameters,][]{he2016deep} on ImageNet, and nanoGPT \citep[$\approx$124\,M parameters][]{karpathy2022nanogpt} on Shakespeare.
We profile performance on an A40 GPU with 40\,GiB of RAM; see \Cref{fig:performance} for the results \& more experimental details.
In both performance metrics, we observe the trend
\begin{equation*}
  \text{KFAC (inverse)}
  \quad\ll\quad
  \text{Fisher/GGN}
  \quad
  <
  \quad
  \text{Hessian}\,.
\end{equation*}
More details in the following paragraphs.

\paragraph{KFAC costs slightly more than a gradient.} Computing KFAC's Kronecker factors costs 1.5-2.5 gradients in terms of time, and the overhead is larger for networks with convolutional layers (ResNet50) that require an additional input unfolding operations.
For both setups, the additional matrix inversion required by the KFAC inverse adds only a barely noticeable time overhead and storing the Kronecker factors increases KFAC's memory consumption by no more than 10-20\,\% compared to the gradient.
Many practical algorithms compute the Kronecker factors infrequently to further reduce costs.
Once the Kronecker factors are computed, applying the curvature approximation to a vector is cheap: we observe run times between 0.1-0.2 gradients.

\paragraph{HVPs match related work.} We find that the HVP performance reflects the findings of \citet{dagreou2024how} who performed an extensive evaluation of HVPs in multiple frameworks and different autodiff modalitites.
For ViT and BERT architectures in PyTorch, they found that HVPs can take up to four to five gradients (we observe 4.5-5.5 gradients) in terms of time, and 3-3.5 times the memory (we observe 3-4.5).
Despite these significantly higher memory demands, note that \texttt{curvlinops} would still allow to split the data into smaller batches, which can reduce the memory consumption back to that similar to a gradient (assuming the peak memory is dominated by stored activations).

\paragraph{FVPs \& GGNVPs are cheaper than HVPs.}
\citet{martens2010deep} found that a matrix-vector product with the GGN (or the Fisher) ``uses about half the memory and runs nearly twice as fast.''
They likely used an autodiff backend different from PyTorch.
While we do not find that GGNVPs are twice as efficient as HVPs, all GGN/Fisher multiplication routines are generally faster than HVPs and consume less memory.

\paragraph{There is potential for further improvements.}
The findings of \citet{dagreou2024how,martens2010deep} suggest that, while \texttt{curvlinops}'s matrix-vector products are of reasonable performance, they can further be accelerated in the future.
Two specific shortcomings are the lack of forward mode AD (all operations are implemented with PyTorch's reverse mode, even JVPs, for maximum operator coverage) and compilation.
As demonstrated by \citet{dagreou2024how}, using these two mechanisms in JAX allows to compute HVPs in roughly 2-4 gradients in time and 2-3 gradients in memory.
We plan to address the forward mode limitation in future releases by switching to the functional API (which according to \citet{dagreou2024how} currently does not support compilation).
Beyond these limitations, we also found other caveats that harm performance.
For nanoGPT, we had to disable PyTorch's efficient attention implementation for certain linear operators as it does not support double backpropagation.
Clearly, this negatively affects performance.

\section{More Curvature Matrices}\label{sec:curvature_matrices_more}
\textbf{Type-II Fisher.}\quad
Integration by parts of the expectation in the type-I Fisher~\eqref{eq:fisher-type-I} leads to the type-II Fisher which highlights the Fisher's connection to the Hessian,
\begin{align}
  \label{eq:fisher-type-II}
 \hspace{-12pt}
  \mF_{\text{II}}(\vtheta)
   & \textstyle\coloneq
  R_{\sD} \;
  \sum_{\vx_n \in \sD}
  {\mathrm{J}_{\vtheta} \vf_n}^{\top}
  \E_{q(\vy \mid \vf_n)}\!
  \left[
  - \nabla^2_{\vf_n} l_n
  \right]
  \mathrm{J}_{\vtheta} \vf_n\,.
\end{align}
In other words, the Fisher can also be written as the negative log likelihood's expected Hessian \emph{\wrt the model's distribution}, whereas the Hessian's expectation in \eqref{eq:hessian} is \emph{\wrt the empirical distribution} implied by the data.
For many common loss functions, $\nabla_{\vf}^2 \log q(\vy \mid \vf)$ does not depend on $\vy$ and the expectation in \eqref{eq:fisher-type-II} can be dropped; then, the Fisher equals the GGN.
Even if this is not possible, we can apply the same Monte-Carlo techniques for estimating the type-II Fisher.
Note however that each sample now requires $\dim(\vf)$ VJPs due to the loss-prediction Hessian's dimension (instead of one vector-Jacobian product in \Cref{eq:fisher-mc}).

\section{Estimating Matrix Properties}\label{sec:matrix-properties}
\begin{figure*}[!p]
  \centering
  \begin{minipage}{0.495\linewidth}
    \centering
    \begin{align*}
      \mA &= \mX + \mZ \mZ^{\top} \in \sR^{2000 \times 2000}
      \\
          &[\mX]_{1,1} = 5, [\mX]_{2,2} = 4, [\mX]_{3,3} = 3
      \\
          &\mZ \in \sR^{2000 \times 2000}, [\mZ]_{i,j} \stackrel{\text{i.i.d}}{\sim} \gN(0, 1)
    \end{align*}
  \end{minipage}
  \hfill
  \begin{minipage}{0.495\linewidth}
    \centering
    \begin{align*}
      \mA &= \frac{1}{1000} \mZ \mZ^{\top} \in \sR^{500 \times 500}
      \\
          &\mZ \in \sR^{500\times 500}, [\mZ]_{i,j} \stackrel{\text{i.i.d}}{\sim} \mathrm{Pareto}(\alpha=1)
    \end{align*}
  \end{minipage}

  \vspace{2ex}
  \textbf{Spectral density estimation}

  \begin{minipage}{0.495\linewidth}
    \centering
    \includegraphics{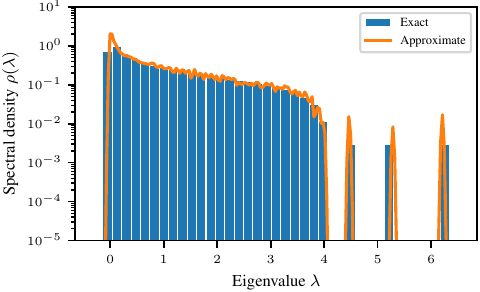}
  \end{minipage}
  \hfill
  \begin{minipage}{0.495\linewidth}
    \centering
    \includegraphics{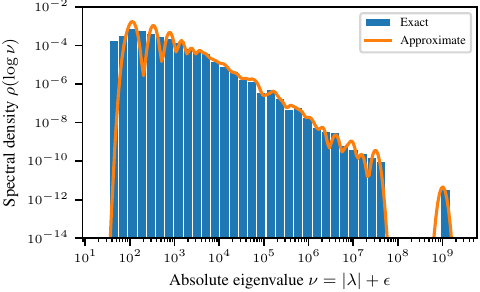}
  \end{minipage}

  \vspace*{-1.5ex}

  \begin{minipage}[c]{0.33\linewidth}
    \centering
    \textbf{Trace estimation}
  \end{minipage}
  \hfill
  \begin{minipage}[c]{0.33\linewidth}
    \centering
    \begin{align*}
      &\mA = \mQ^\top \diag(\vlambda) \mQ \in \sR^{1000 \times 1000}\,,
      \\
      &\mQ \mQ^{\top} = \mI\,, \quad
      [\vlambda]_i = c^{-i}
    \end{align*}
  \end{minipage}
  \hfill
  \begin{minipage}[c]{0.33\linewidth}
    \centering
    \textbf{Diagonal estimation}
  \end{minipage}

  \begin{minipage}{0.495\linewidth}
    \centering
    \includegraphics{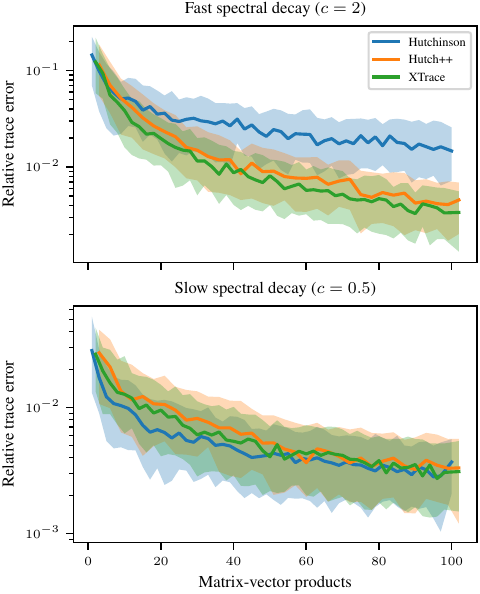}
  \end{minipage}
  \hfill
  \begin{minipage}{0.495\linewidth}
    \centering
    \includegraphics{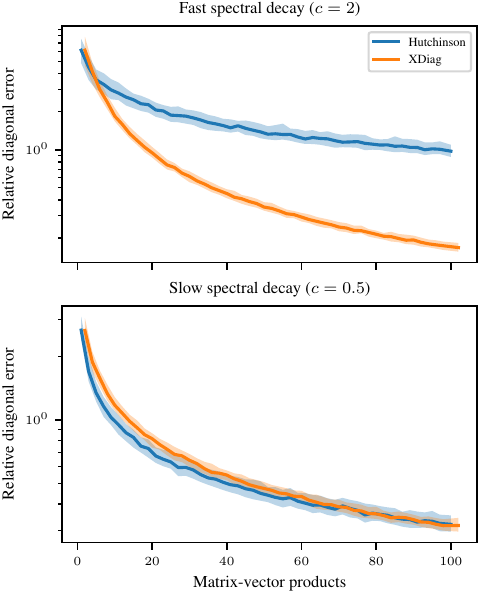}
  \end{minipage}

  \caption{\textbf{Estimating linear operator properties with \texttt{curvlinops}.}
    We implement various estimation algorithms from the literature and evaluate them on toy problems.
    \emph{Top:} Spectral density estimation with the algorithms and toy matrices from \citet{papyan2020prevalence}.
    The left panel estimates a spectral density, the right panel the spectral density of the matrix logarithm $\log(|\mA| + \epsilon \mI)$ with $\epsilon= 10^{-5}$.
    Code to reproduce these figures is \href{\linkToDocs/en/latest/basic_usage/example_verification_spectral_density.html}{here}.
    \emph{Bottom:} Comparison of trace and diagonal estimators~\cite{girard1989montecarlo,hutchinson1989stochastic,epperly2024xtrace,meyer2020hutch} for matrices whose spectrum follows a power law ($\mQ$ is obtained from the QR decomposition of a random Gaussian matrix).
    Solid lines are medians, error bars are 25- and 75-percentiles over 200 runs.
    For traces, we use the relative error $\nicefrac{|\hat{t} - t|}{|t|}$ where $\hat{t}$ estimates the true trace $t$.
    For diagonals, we report the relative error $\nicefrac{\max_i |a_i - \hat{a}_i|}{\max_j|a_j|}$ where $\hat{\va}$ approximates the true diagonal $\va$.
    On matrices with fast spectral decay, estimation techniques based on variance reduction improve over vanilla Hutchinson estimators.
  }\label{fig:matrix-properties}
\end{figure*}

To verify the correctness of our implemented matrix property estimation techniques, we perform sanity checks on toy problems from the original works.
\cref{fig:matrix-properties} visualizes our findings.
See \cref{sec:interoperability} for a more detailed discussion.

\end{document}